\pdfoutput=1
\documentclass[sigconf]{acmart}

\usepackage{booktabs}
\usepackage{multirow}
\usepackage{graphicx}
\usepackage{amsmath}
\usepackage{microtype}
\acmConference[PEARC '26]{Practice and Experience in Advanced Research Computing}{July 26--30, 2026}{Minneapolis, Minnesota, USA}
\acmYear{2026}
\copyrightyear{2026}
\setcopyright{cc}
\setcctype{by}
\acmDOI{}
\acmISBN{}

\begin{document}

\title{LLMs or Naive Bayes? Old Gems or New Ways}

\author{Mohammad Firas Sada}
\orcid{0009-0006-6045-2940}
\email{mfsada@ucsd.edu}
\affiliation{%
  \institution{San Diego Supercomputer Center, University of California, San Diego}
  \city{La Jolla}\state{CA}\country{USA}
}
\author{John Graham}
\orcid{0000-0002-2139-5617}
\email{jjgraham@ucsd.edu}
\affiliation{%
  \institution{San Diego Supercomputer Center, University of California, San Diego}
  \city{La Jolla}\state{CA}\country{USA}
}
\author{Mahidhar Tatineni}
\orcid{0009-0003-0709-090X}
\email{mahidhar@sdsc.edu}
\affiliation{%
  \institution{San Diego Supercomputer Center, University of California, San Diego}
  \city{La Jolla}\state{CA}\country{USA}
}
\author{Dmitry Mishin}
\orcid{0000-0003-1125-448X}
\email{dimm@lbl.gov}
\affiliation{%
  \institution{Lawrence Berkeley National Laboratory}
  \streetaddress{1 Cyclotron Road}
  \city{Berkeley}\state{CA}\postcode{94720}\country{USA}
}
\author{Seungmin Kim}
\orcid{0000-0001-8052-723X}
\email{ehf@yonsei.ac.kr}
\affiliation{%
  \institution{Yonsei University College of Medicine}
  \city{Seoul}\country{Republic of Korea}
}
\author{Frank W\"{u}rthwein}
\orcid{0000-0001-5912-6124}
\email{fkw@physics.ucsd.edu}
\affiliation{%
  \institution{San Diego Supercomputer Center, University of California, San Diego}
  \city{La Jolla}\state{CA}\country{USA}
}

\begin{abstract}
Large language models (LLMs) prompt a recurring question in research computing: should classical methods like Naive Bayes (NB) be retired? We benchmark Complement Naive Bayes against zero-shot and few-shot LLMs spanning four model families and a $37\times$ range in scale (27B to a 1T-parameter mixture-of-experts) across text classification tasks. LLMs dominate only in zero-data regimes (98.0\% vs 88.2\% on Amazon Polarity sentiment), and even that win is contamination-prone: on a low-contamination sentiment task NB \emph{beats} the zero-shot LLM (81.7\% vs 73.0\%). However, once labeled data is available (e.g., AG News), NB reaches 89.1\% accuracy, statistically indistinguishable from the zero-shot 27B LLM (89.0\%) and \emph{better} than the 397B frontier model (84.8\%), at thousands of samples/sec on a commodity CPU. Fine-tuned DistilBERT reaches 90.6\% but at far lower throughput than NB at batch size~1 (Table~\ref{tab:agnews}). Our measured GPU throughput analysis shows small-LLM batched inference is 40--486$\times$ slower than NB CPU inference (the multiplier depends strongly on the host CPU), exposing a structural gap bounded by memory bandwidth, with roughly two orders of magnitude lower energy per sample. For resource-constrained HPC practitioners performing text classification with labeled data, NB remains the optimal choice. We show the decision line is task-dependent (NB reaches LLM parity around $N\sim10^4$ labels for topic classification, while zero-data sentiment favors the LLM at all $N$ tested) and provide a Kubernetes Helm operator that automates model selection using configurable thresholds and verifiable Prometheus metrics.
\end{abstract}



\maketitle

\section{Introduction}
\label{sec:intro}

The rise of large language models (LLMs) such as GPT-4 and open-source families like Qwen3 has reshaped natural language processing \citep{brown2020gpt3,ouyang2022instructgpt}, raising a concrete question for practitioners who routinely classify text at scale: \emph{should we retire classical probabilistic classifiers like Naive Bayes?} The question carries real resource implications: researchers on shared HPC clusters face steep energy and financial costs \citep{strubell2019energy,schwartz2020green} when deploying large models, whereas Complement Naive Bayes (CNB) \citep{rennie2003tackling} with TF-IDF bigrams \citep{wang2012baselines} trains in under one second on a commodity CPU and classifies at sub-millisecond speeds per sample (Section \ref{sec:throughput}).

Concurrent work on serving LLMs in HPC clusters~\citep{sada2025serving} shows modern accelerators improve energy efficiency but cannot match the per-sample efficiency of classical methods. We benchmark CNB against zero/few-shot LLMs spanning 27B to 1T parameters: LLMs hold a structural advantage only when labeled data is absent, while with labels NB matches or beats zero-shot LLMs at a measured 40--486$\times$ throughput advantage driven by memory-bandwidth saturation, across batch sizes and two accelerators. For the many scientific text-classification pipelines that already possess labeled data, moving to an LLM adds unnecessary complexity and cost. We then characterize the task-dependent decision line where NB becomes superior and operationalize it with a configurable Kubernetes Helm operator that exports Prometheus metrics for verifiability.

\section{Related Work}
\label{sec:related}

\citet{rennie2003tackling} introduced Complement Naive Bayes; TF-IDF linear models remain competitive \citep{wang2012baselines}; fine-tuned small models often beat zero-shot LLMs when labels exist \citep{bucher2024finetuned,schucher2023fewshot}. In the low-label regime, prompt-free label-efficient methods such as SetFit \citep{tunstall2022setfit} and parameter-efficient few-shot tuning (T-Few) \citep{liu2022fewshot} offer an alternative to in-context learning, while recent zero-shot benchmarks compare cross-encoders, rerankers, and LLMs \citep{aarab2026btzsc}. We instead characterize the low-label regime directly via training-size curves (Fig.~\ref{fig:ncurve}) that locate where NB reaches LLM parity. Energy costs of large-scale inference are well documented \citep{strubell2019energy,schwartz2020green,sada2025serving}.

\paragraph{Systems and platforms.}
Our benchmarks execute on the same class of federated scientific Kubernetes
infrastructure as the National Research Platform (NRP)~\cite{weitzel2025national}.
Complementary work from our group has compared LLM serving on Qualcomm Cloud AI and
NVIDIA GPUs in shared HPC clusters~\cite{sada2025serving}; studied real-time
P4-programmable FPGA networking for in-network machine learning~\cite{sada2025fpga};
and presented SCinet NRE demonstrations spanning live NRP telemetry and
P4/FPGA/DPU testbeds~\cite{sadasc25telemetry,sadasc25p4}.

\section{Methodology}
\label{sec:methodology}

\subsection{Models}
We evaluate models spanning classical to large-scale.
\textbf{Classical baselines}: Complement Naive Bayes (CNB)~\citep{rennie2003tackling} with TF-IDF bigrams~\citep{wang2012baselines} and Logistic Regression (LR), both CPU-only.
\textbf{Fine-tuned transformer}: DistilBERT (66M)~\citep{sanh2019distilbert}, fine-tuned with AdamW (lr $3{\times}10^{-5}$, batch size 32).
\textbf{Generative LLMs}: Qwen3.6-27B (served checkpoint \texttt{Qwen/\allowbreak Qwen3.6-27B}) in zero-shot and few-shot ($k{=}5$) settings, and the Qwen3.5 (397B) frontier model (\texttt{Qwen/\allowbreak Qwen3.5-\allowbreak 397B-\allowbreak A17B-\allowbreak FP8}, a 397B mixture-of-experts with 17B active parameters) in zero-shot, served on the managed NRP endpoint~\citep{weitzel2025national}.
\textbf{Discriminative LLM}: a LoRA fine-tuned Qwen3-8B sequence classifier~\citep{hu2022lora} that predicts in a single forward pass, included so a large model is compared on equal (discriminative) footing rather than via autoregressive generation.
\textbf{Throughput probe}: OLMo-2-1B (\texttt{allenai/\allowbreak OLMo-2-\allowbreak 0425-1B-\allowbreak Instruct}, 1.48B)~\citep{groeneveld2024olmo} on GPU.
All generative inference uses temperature~0.

\subsection{Datasets}
We use three benchmarks with stratified splits (seed 42).  
\textbf{Amazon Polarity}~\citep{zhang2015character}: binary sentiment, 10K/2K train/test; high pre-training overlap for LLMs.  
\textbf{AG News}~\citep{zhang2015character}: four-class topic classification, 10K/2K; clean and standard.  
\textbf{20 Newsgroups}: 20-class topic classification, 11{,}314/7{,}532; headers/footers removed (``hard'' setting). Cross-posting introduces label noise favoring distributional methods.
We cap Amazon and AG~News training at 10K to hold the labeled-data budget constant across datasets (20~Newsgroups is already $\sim$11K) and because the per-sample LLM evaluation cost makes larger comparisons impractical; the training-size curves (Fig.~\ref{fig:ncurve}) show NB and DistilBERT accuracy already saturating well before 10K, so a larger subsample would not change the decision-line conclusion.

\subsection{Evaluation and Infrastructure}
We report accuracy, weighted F1, McNemar tests and paired-bootstrap CIs on per-sample predictions. Throughput uses pre-tokenized samples across batch sizes 1--128 on NVIDIA RTX~3090 (24\,GB) and Tesla V100-SXM2-32GB, with GPU board power read from NVML. CPU baselines run on dual Intel Xeon Gold 6248R (48 cores) with NUMA affinity. Experiments run as Kubernetes batch jobs on the NRP Nautilus cluster~\citep{weitzel2025national}. This study's $\approx$15 GPU-hours of self-hosted compute, at the measured board power and U.S.\ grid intensity (367\,gCO$_2$e/kWh)~\citep{eia2024us}, give an operational footprint of order $10^0$\,kgCO$_2$e under SCI~\citep{greensoftware2025sci}, negligible beside the $\approx$2{,}159\,kgCO$_2$e embodied carbon of a single V100 GPU server~\citep{ji2024scarif}.

\subsection{LLM Endpoint Details}
The Qwen3.6-27B and Qwen3.5 (397B) models are served with vLLM~\citep{vllm0171} (continuous batching) on the managed NRP endpoint, which we use for all accuracy measurements. Controlled throughput and power measurement (Table~\ref{tab:throughput}) instead uses models we host ourselves on dedicated GPUs, namely OLMo-2-1B as a small generative probe and DistilBERT, so that GPU board power (NVML) is attributable to a single co-located workload. Decoding uses temperature~0 and max sequence length 8{,}192.

\subsection{Memory-Bandwidth Analysis}
The batch-1 inference lower bound is $t_{\text{min}} = 2P / B_{\text{mem}}$, where $P$ is the parameter count and $B_{\text{mem}}$ is memory bandwidth. For OLMo-2-1B (1.48\,B params, $\approx$2.96\,GB in FP16; RTX~3090 at 936\,GB/s), $t_{\text{min}} \approx 3.2$\,ms/sample ($\approx$316\,samples/s). Our measured peak of 200\,samples/s reaches $\approx$63\% of this bound, the remainder lost to attention/KV and kernel-launch overhead; batching past size~8 does not raise throughput because the decode step remains weight-streaming-bound rather than compute-bound. The NB TF-IDF model fits in CPU cache and approaches peak sparse-BLAS throughput.

\subsection{Reproducibility}
\label{sec:artifacts}
We fix splits (seed 42), tokenizer settings, and vLLM launch flags; jobs use namespace-scoped quotas consistent with other NRP tenants~\citep{weitzel2025national}. A public artifact bundle (scripts, manifests, and configs to reproduce Tables~\ref{tab:amazon}--\ref{tab:throughput} with explicit NUMA/CUDA pinning) is available at \url{https://github.com/groundsada/llms-or-naive-bayes}.

\section{Results}
\label{sec:results}

\subsection{Amazon Polarity: Binary Sentiment}

On Amazon Polarity (Table~\ref{tab:amazon}), sentiment is easy for LLMs given heavy
pre-training overlap: zero-shot Qwen3.6-27B reaches 97.8\% and few-shot 98.0\%, versus 88.2\%
for CNB, a gap that is statistically significant (McNemar $p<0.05$; the paired-bootstrap CI on
the accuracy difference excludes~0). The Qwen3.5 (397B) \emph{frontier} model does not improve on
the 27B (98.0\%), indicating the ceiling here is the task, not model scale. A discriminative
fine-tuned Qwen3-8B reaches 97.2\%, close to the generative LLMs but at a single forward pass
(91\,ms vs.\ $\sim$13\,s). LR is statistically indistinguishable from CNB. Crucially, CNB
classifies at 0.06\,ms/sample, roughly five orders of magnitude faster than the generative LLMs.
We flag this zero-shot win as \emph{contamination-prone} (Amazon Polarity is a long-public
benchmark with high pre-training overlap) and test it on a low-contamination sentiment task
in \S\ref{sec:contam}, where the LLM advantage reverses.

\begin{table}[h]
\centering
\footnotesize
\caption{Amazon Polarity (2K test; LLM rows on 500). Latency at batch size~1; generative-LLM
latency includes reasoning tokens.}
\label{tab:amazon}
\begin{tabular}{lccr}
\toprule
Model & Acc & F1 & ms/sample \\
\midrule
CNB & .882 & .882 & \textbf{0.06} \\
LR & .887 & .890 & \textbf{0.08} \\
DistilBERT (fine-tuned) & .935 & .935 & 5.4 \\
Qwen3-8B (fine-tuned, discriminative) & .972 & .972 & 91 \\
Qwen3.6-27B zero-shot  & .978 & .978 & 12{,}968 \\
Qwen3.6-27B few-shot $k{=}5$ & .980 & .980 & 8{,}684 \\
Qwen3.5 (397B) zero-shot & \textbf{.980} & \textbf{.980} & 13{,}734 \\
\bottomrule
\end{tabular}
\end{table}

\subsection{AG News: Clean Multi-Class}

On AG News (Table~\ref{tab:agnews}), once labels are available CNB (89.1\%) is statistically
\emph{indistinguishable} from the zero-shot 27B LLM (89.0\%; McNemar $p{=}0.07$) and from
few-shot (88.4\%), at 0.03\,ms/sample versus $\sim$13\,s. Strikingly, the 397B frontier model
is \emph{worse} (84.8\%): the reasoning-tuned model over-generates on a simple four-way topic
task, so scale does not help here. The fine-tuned models lead on accuracy (DistilBERT 90.6\%
and the discriminative Qwen3-8B 93.3\%), but the latter still costs 42\,ms/sample (a single
forward pass), over 1{,}000$\times$ slower than CNB.

\begin{table}[h]
\centering
\footnotesize
\caption{AG News (2K test; LLM rows on 500). Latency at batch size~1.}
\label{tab:agnews}
\begin{tabular}{lccr}
\toprule
Model & Acc & F1 & ms/sample \\
\midrule
CNB & .891 & .890 & \textbf{0.03} \\
LR & .884 & .883 & \textbf{0.04} \\
DistilBERT (fine-tuned) & .906 & .905 & 5.4 \\
Qwen3-8B (fine-tuned, discriminative) & \textbf{.933} & \textbf{.933} & 42 \\
Qwen3.6-27B zero-shot  & .890 & .889 & 13{,}603 \\
Qwen3.6-27B few-shot $k{=}5$ & .884 & .883 & 10{,}274 \\
Qwen3.5 (397B) zero-shot & .848 & .846 & 17{,}143 \\
\bottomrule
\end{tabular}
\end{table}

\subsection{20 Newsgroups: Noisy Multi-Class}

On 20~Newsgroups (Table~\ref{tab:20news}), the true 20-class topic task with headers/footers
removed, CNB (71.2\%) is statistically \emph{indistinguishable} from the zero-shot 27B LLM
(71.8\%; McNemar not significant), and few-shot edges ahead (73.8\%). The 397B frontier model
is again \emph{worse} (66.4\%), significantly below both CNB and the 27B (McNemar $p<0.05$):
the reasoning-tuned model is poorly calibrated for fine-grained many-way topic labels.
DistilBERT trails NB here (67.2\%); with cross-posting label noise and only $\sim$565 examples
per class, the fine-tuned encoder fails to overtake the sparse bag-of-words model at \emph{any}
training size (Fig.~\ref{fig:ncurve}). Only the discriminative fine-tuned Qwen3-8B leads on
accuracy (76.3\%), but at 97\,ms/sample, roughly 700$\times$ slower than CNB, which classifies
at 0.14\,ms/sample (five orders of magnitude faster than the generative LLMs).

\begin{table}[h]
\centering
\footnotesize
\caption{20 Newsgroups (true 20-class; 7,532 test, LLM rows on 500). Latency at batch size~1;
generative-LLM latency includes reasoning tokens. DistilBERT at its largest training size.}
\label{tab:20news}
\begin{tabular}{lccr}
\toprule
Model & Acc & F1 & ms/sample \\
\midrule
CNB & .712 & .706 & \textbf{0.14} \\
LR & .673 & .666 & \textbf{0.15} \\
DistilBERT (fine-tuned) & .672 & .664 & 5.4 \\
Qwen3-8B (fine-tuned, discriminative) & \textbf{.763} & \textbf{.763} & 97 \\
Qwen3.6-27B zero-shot  & .718 & .726 & 19{,}226 \\
Qwen3.6-27B few-shot $k{=}5$ & .738 & .742 & 17{,}407 \\
Qwen3.5 (397B) zero-shot & .664 & .703 & 32{,}212 \\
\bottomrule
\end{tabular}
\end{table}

\subsection{Cross-Family and Scale Generalization}
\label{sec:crossfamily}
A natural objection is that our decision line is an artifact of the Qwen family, or that
a \emph{larger} model would tip the trade-off. We test both by evaluating three additional
\emph{non-Qwen} models under the identical protocol (zero-shot, 500-sample LLM slice,
temperature~0): gpt-oss-120B (OpenAI, MoE), GLM-5.2 (Zhipu, 744B MoE), and Kimi-K2 (1T MoE).
Table~\ref{tab:crossfamily} places these alongside the Qwen 27B and 397B references; together
the evaluated models span four families and a $37\times$ range in scale (27B--1T). The same regime holds across every family and
scale: all models lead decisively only on zero-data Amazon sentiment ($.974$--$.980$,
essentially flat from 120B to 1T), while on the label-rich topic tasks they cluster around or
\emph{below} Complement NB. Crucially, accuracy is not monotone in scale: on AG~News the 27B
Qwen ($.890$) matches CNB and \emph{beats} the 397B, 120B, and 744B models, and on
20~Newsgroups the 397B is the \emph{worst} of all, so parameter count is not the lever that
moves the decision line. The trade-off is governed by the task and by label availability, not
by model lineage or scale.

\begin{table}[h]
\centering
\footnotesize
\caption{Cross-family/scale check: zero-shot accuracy of three non-Qwen models, spanning
120B--1T parameters, vs.\ the CNB and Qwen references, on the shared 500-sample LLM slice
(CNB on full test). The label-rich topic tasks neutralize the LLM advantage across every
family \emph{and} scale tested; larger models do not help once labels are present.}
\label{tab:crossfamily}
\begin{tabular}{llccc}
\toprule
Model & Params & Amazon & AG News & 20NG \\
\midrule
Complement NB        & --         & .882 & \textbf{.891} & .712 \\
\midrule
Qwen3.6-27B          & 27B        & .978 & .890 & .718 \\
gpt-oss-120B         & 120B       & .974 & .832 & .692 \\
Qwen3.5 (frontier)   & 397B-A17B  & \textbf{.980} & .848 & .664 \\
GLM-5.2              & 744B       & \textbf{.980} & .882 & .726 \\
Kimi-K2              & 1T         & .978 & .870 & \textbf{.764} \\
\bottomrule
\end{tabular}
\end{table}

\subsection{GPU Throughput Analysis}
\label{sec:throughput}

Table~\ref{tab:throughput} reports measured throughput with the NB CPU baseline and the GPU
models benchmarked on the \emph{same} node. On the RTX~3090 node, NB classifies at
7{,}979~samp/s while OLMo-2-1B peaks at 200~samp/s (batch~8) and DistilBERT at
1{,}678~samp/s, a 40$\times$ and 4.8$\times$ NB advantage, respectively. On the V100 node
(faster host CPU), NB reaches 21{,}587~samp/s while OLMo peaks at just 44~samp/s, a
486$\times$ gap. GPU throughput saturates by batch~8 and does not recover, consistent with a
memory-bandwidth-bound regime in which each forward pass streams the full weight matrix from
VRAM. The exact multiplier therefore depends strongly on the host CPU (40--486$\times$ here),
but in every configuration the sparse linear model is one to nearly three orders of magnitude
faster than autoregressive GPU inference. These multipliers use OLMo-2-1B, the
\emph{smallest} generative model; the 27B and 397B models that produced our accuracy rows are
far larger and, under the same memory-bandwidth bound, necessarily slower per sample (their
managed-endpoint latency is $\sim$13--32\,s/sample, Tables~\ref{tab:amazon}--\ref{tab:20news}).
The reported gap is therefore a \emph{conservative lower bound} on the throughput penalty of
the LLMs that actually achieve the headline accuracies.

\begin{table}[h]
\centering
\footnotesize
\caption{Measured throughput and GPU board power (NVML), NB (CPU) vs.\ GPU models on the
\emph{same} node, pre-tokenized, each point sampled for $\ge$5\,s. ``vs.\ NB'' = NB$\div$model
throughput. GPU peaks are at batch size~8 (OLMo) / 8--32 (DistilBERT).}
\label{tab:throughput}
\begin{tabular}{llrrr}
\toprule
Node & Model & Samp/s & GPU\,W & vs.\ NB \\
\midrule
\multirow{3}{*}{RTX 3090} & \textbf{NB (CPU)} & \textbf{7{,}979} & n/a & $1\times$ \\
 & DistilBERT & 1{,}678 & 349 & $4.8\times$ \\
 & OLMo-2-1B & 200 & 341 & $40\times$ \\
\midrule
\multirow{3}{*}{V100-32GB} & \textbf{NB (CPU)} & \textbf{21{,}587} & n/a & $1\times$ \\
 & DistilBERT & 1{,}343 & n/a & $16\times$ \\
 & OLMo-2-1B & 44 & 272 & $486\times$ \\
\bottomrule
\end{tabular}
\end{table}

\section{Discussion}
\label{sec:discussion}

\begin{figure*}[t]
\centering
\includegraphics[width=\linewidth]{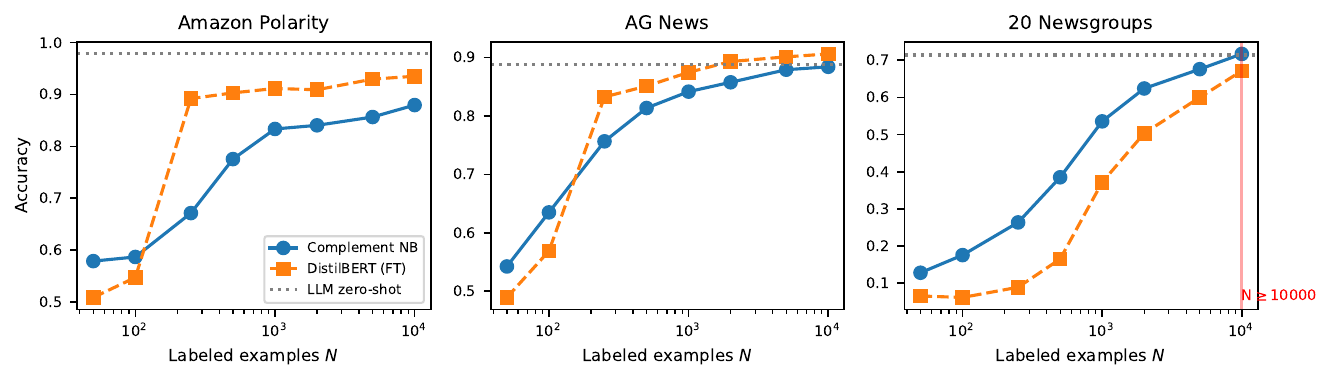}
\Description{Three line plots of classification accuracy versus the number of labeled
training examples on a logarithmic horizontal axis, one panel each for Amazon Polarity, AG
News, and 20 Newsgroups. Each panel plots Complement Naive Bayes and fine-tuned DistilBERT as
rising curves and the zero-shot 27B large language model as a flat dotted reference line. Naive
Bayes meets the language-model line only near ten thousand examples on 20 Newsgroups and AG
News, and never on Amazon within the tested range.}
\caption{Decision line: accuracy of Complement NB and fine-tuned DistilBERT versus the number
of labeled training examples $N$ (log scale), with the zero-shot 27B LLM as a constant
reference (dotted). The labeled break-even $N$ (smallest $N$ at which NB reaches the zero-shot
LLM) is task-dependent: NB matches the LLM only at $N\!\approx\!10^4$ on 20~Newsgroups,
approaches it by $N\!\approx\!10^4$ on AG News, and never reaches it on Amazon sentiment within
$N\le10^4$. There is no single universal threshold.}
\label{fig:ncurve}
\end{figure*}

\paragraph{The decision line is task-dependent, not a single threshold.}
Figure~\ref{fig:ncurve} traces accuracy against the number of labeled examples $N$. Contrary
to a fixed ``$N\!\ge\!500$'' rule, the labeled break-even where NB matches the zero-shot 27B
varies sharply by task: $N\!\approx\!10^4$ on 20~Newsgroups and AG~News, but never within
$N\le10^4$ on Amazon sentiment. DistilBERT overtakes NB at moderate $N$ on the two
binary/clean-topic tasks ($N\!\approx\!250$) but trails NB at \emph{every} $N$ on noisy
20-class 20~Newsgroups. Critically, the cost axis does not move with $N$: NB retains its
40--486$\times$ throughput and $\sim$two-orders-of-magnitude energy advantage regardless of
label count, so whenever NB is within an application's accuracy tolerance it is the economical
choice, which is why we operationalize the boundary as a configurable policy, not a fixed constant.

Our results expose a structural GPU throughput gap driven by memory-bandwidth limits: each forward pass streams full weights from VRAM, while NB's sparse TF-IDF fits in CPU cache. At its 200~samp/s peak, OLMo-2-1B draws a \emph{measured} 341~W (NVML), i.e.\ 1.70~J/sample. CPU package energy was not exposed via RAPL inside our containers, so we bound NB conservatively by its host-CPU package TDP ($\sim$150~W): at 7{,}979~samp/s this is $\approx$0.02~J/sample, roughly two orders of magnitude below the LLM. Even under this conservative bound, classifying $10^7$ documents differs by order $10^2$--$10^3\times$ in energy, a meaningful operational saving consistent with Green AI~\citep{schwartz2020green,sada2025serving}. We avoid extrapolating to annual fleet-scale CO$_2$e, which depends on workload mix and grid intensity beyond our measurements.

\begin{figure}[t]
\centering
\includegraphics[width=0.82\linewidth]{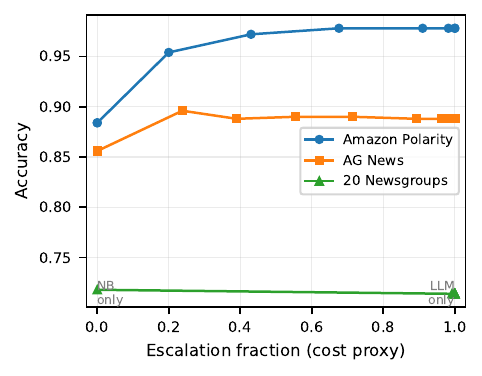}
\Description{Line plot of classification accuracy versus the escalation fraction (the share
of samples routed from Naive Bayes to the LLM, a proxy for cost) for the three datasets. The
left endpoint is Naive Bayes alone and the right endpoint is the LLM alone. Amazon Polarity
rises smoothly from 0.88 to 0.98 as escalation increases; AG News peaks near 0.90 at about a
quarter escalation, above both endpoints; 20 Newsgroups stays flat because Naive Bayes is
already best.}
\caption{Cascade accuracy--cost frontier. Routing only NB's low-confidence samples (threshold
$\tau$) to the LLM traces a tunable operating point between NB-only (left, free) and LLM-only
(right, full cost). On AG~News a small escalation \emph{exceeds} both endpoints; on
20~Newsgroups NB alone is already best, so escalation only adds cost.}
\label{fig:cascade}
\end{figure}

\paragraph{Hybrid NB$\to$LLM cascade.}
Rather than choosing NB \emph{or} the LLM, we can route only NB's low-confidence predictions
(max class probability below a threshold $\tau$) to the LLM. Sweeping $\tau$ traces an
accuracy--cost frontier (Fig.~\ref{fig:cascade}), where the escalation fraction is a direct
proxy for cost. On Amazon,
escalating 43\% (68\%) of samples reaches 97.2\% (97.8\%), matching the standalone LLM
(97.8\%) while keeping more than half the workload on the CPU. On AG News, a threshold that
escalates only 24\% of samples reaches 89.6\%, \emph{exceeding} both standalone NB (85.6\%) and
the zero-shot LLM (88.8\%). On 20~Newsgroups NB alone is already best (71.8\% vs.\ 71.4\%), so
escalation only adds cost. The cascade thus turns the accuracy--throughput trade-off into a
single tunable operating point.

\begin{figure}[t]
\centering
\includegraphics[width=0.92\linewidth]{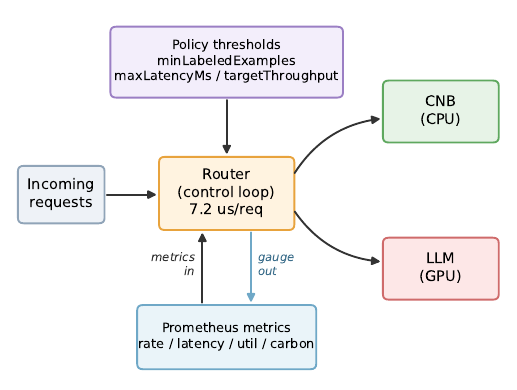}
\Description{Block diagram of the Helm operator. Incoming requests enter a central
router (control loop) with a measured 7.2 microsecond per-request overhead. The router
is configured by policy thresholds (minLabeledExamples, maxLatencyMs, targetThroughput)
and reads live Prometheus metrics (rate, latency, utilization, carbon); it routes each
request to either the Complement Naive Bayes CPU endpoint or the LLM GPU endpoint, and
exports the active choice as a one-hot selected-model gauge back to Prometheus.}
\caption{Helm operator architecture. A policy- and metrics-driven router dispatches each
request to the CNB (CPU) or LLM (GPU) endpoint and exports the active choice as a one-hot
\texttt{selected\_model} gauge (plus per-model counters and decision/inference histograms)
to Prometheus. The control loop adds a measured 7.2\,$\mu$s per request.}
\label{fig:operator}
\end{figure}

\paragraph{Operationalizing the Decision Boundary: A Helm Operator}
\label{sec:operator}
Because the break-even point is task-dependent rather than a single universal threshold, we expose it as a \emph{configurable} policy rather than hard-coding it. We implement a Helm operator (Fig.~\ref{fig:operator}) that automates model selection: it exposes thresholds (e.g., \texttt{minLabeledExamples}, \texttt{maxLatencyMs}, \texttt{targetThroughput}) that an operator sets from the relevant task's decision line, and continuously ingests Prometheus metrics from NB and LLM endpoints (rate, latency, utilization, carbon). Requests are routed to the cheaper model whenever it satisfies the accuracy/latency policy, with the active choice exported as a one-hot \texttt{selected\_model} gauge plus per-model counters and latency histograms, so the routing decision is auditable in Prometheus without string-valued metrics. The routing layer is effectively free: a server-side microbenchmark measures a mean decision overhead of \textbf{7.2~$\mu$s} per request (211 requests), negligible against either model's inference time. The accuracy--cost frontier of Fig.~\ref{fig:cascade} is precisely the operating curve the operator selects along, so the policy is not merely plumbing: choosing the threshold moves the deployment to a measured, better accuracy/cost point (e.g., the AG~News knee that beats both NB-only and LLM-only). Concretely, with \texttt{minLabeledExamples} set to a task's measured break-even (Fig.~\ref{fig:ncurve}), the routed model flips from the LLM to CNB exactly as the available label count crosses that point: the decision boundary made operational. This turns our findings into a deployable control loop for Kubernetes environments.

\paragraph{Reproducibility and deployment context.}
Artifacts are summarized in \S\ref{sec:artifacts}. Jobs run in standard NRP namespaces with pinned images on shared infrastructure used for prior accelerator and networking studies~\cite{sada2025serving,sada2025fpga}. The reported CPU--GPU gaps are thus conservative under mixed workloads and shared power budgets~\cite{weitzel2025national}.

\textbf{Practical guidance.} Use zero-shot LLMs when no labels exist; once enough labels are available to reach NB--LLM parity for the task (order $10^4$ for topic classification in our experiments), switch to CNB for similar accuracy at one-to-nearly-three orders of magnitude lower cost. Fine-tune DistilBERT/RoBERTa only if a 1--3 pp gain justifies GPU use. Streaming workloads favor NB and batching does not close the gap, so for high-volume pipelines with labeled data NB minimizes both latency and cost; the Helm operator enforces the chosen policy automatically.

\textbf{Limitations.} English benchmarks only; biomedical or multilingual settings may differ. Our generative LLMs span four model families (Qwen, gpt-oss, GLM, Kimi) and a $37\times$ range in scale (27B--1T); none improved accuracy over NB once labels were present (Table~\ref{tab:crossfamily}), so the decision line is not an artifact of one lineage or of insufficient scale. Specialized or much larger models could still shift the trade-off on harder, long-context tasks we did not test.

\paragraph{Contamination control.}
\label{sec:contam}
The Amazon zero-shot headline sits on a long-public benchmark with high pre-training overlap,
so its 98\% may partly reflect memorization. Re-running the identical protocol on a
domain-specific sentiment task that is not a canonical benchmark and whose labels resist
surface-word shortcuts (financial-news sentiment,
bearish/bullish/neutral)\footnote{\url{https://huggingface.co/datasets/zeroshot/twitter-financial-news-sentiment}}
reverses the LLM advantage: Complement NB reaches 81.7\%, significantly above the zero-shot
27B (73.0\%; McNemar $p{=}0.002$), few-shot (70.6\%; $p{<}10^{-3}$), and the 397B frontier
(63.6\%; $p{<}10^{-8}$). Once contamination risk is reduced, NB beats the zero-shot LLM even in
the zero-label sentiment regime it was expected to dominate, and, consistent with
\S\ref{sec:crossfamily}, scale again hurts. The conclusion is robust to the dataset's exact
provenance: any residual contamination would \emph{inflate} the LLM scores, yet they fall well
\emph{below} NB, so the direction of the effect cannot be a memorization artifact.

\paragraph{Threats to validity.}
Benchmarks favor bag-of-words; long-range semantic tasks may widen transformer gains despite throughput costs. Qwen endpoints use a fixed vLLM release~\citep{vllm0171}; kernel or default changes may shift latencies without altering the order-of-magnitude gap. We assume warm caches and steady clocks; thermal throttling would only further reduce LLM throughput.
\section{Conclusion}
\label{sec:conclusion}

Naive Bayes is not dead. For text classification tasks with labeled data, Complement Naive Bayes with TF-IDF bigrams remains optimal for most HPC practitioners: $\approx$89\% accuracy at thousands of samples/sec on CPU, with sub-second training and zero GPU allocation. No batch size of GPU-accelerated transformer inference approaches this throughput; the measured gap is 40--486$\times$ (depending on the host CPU) and is bounded by memory bandwidth. Notably, even a 397B frontier model does not beat NB once labels are present. LLMs excel only in zero-shot settings without labeled data. For HPC practitioners, the choice is clear: use NB when labeled data exists; use LLMs only when it does not. We have characterized the task-dependent decision line (NB reaches LLM parity around $N\sim10^4$ labels for topic classification) and operationalized it as a configurable Helm operator with Prometheus-based verifiability and a measured 7.2~$\mu$s routing overhead. As HPC systems increasingly emphasize energy efficiency and resource sharing, classical methods like NB deserve renewed attention. Future work should explore hybrid pipelines that use NB for rapid filtering and LLMs only for ambiguous cases.

On the same NRP substrate that runs our endpoints and related accelerator and dataplane
studies~\cite{sada2025serving,sada2025fpga,sadasc25p4,sadasc25telemetry}, classical
baselines free scarce GPU and network capacity for workloads that genuinely require them.

\begin{acks}
We thank the open-source AI community for making the models and software used in this study publicly available. We also thank Derek Weitzel, Ashton Graves, Sam Albin, Huijun Zhu, and Daniel Diaz at the University of Nebraska--Lincoln and UC San Diego for maintaining the National Research Platform (NRP) infrastructure~\citep{weitzel2025national}, which hosted all LLM benchmarks. This work was supported in part by National Science Foundation (NSF) awards CNS-1730158, ACI-1540112, ACI-1541349, OAC-1826967, OAC-2112167, CNS-2100237, and CNS-2120019. Portions of this paper were edited using LLMs hosted on NRP~\citep{weitzel2025national}.

Seungmin Kim was supported by the Seok-San Yonsei Medical Scientist Training Program (MSTP) Song Yong-Sang Scholarship and a Student Research Bursary from the Song-Dang Institute for Cancer Research at Yonsei University College of Medicine, the MD-PhD/Medical Scientist Training Program (MSTP) through the Korea Health Industry Development Institute (KHIDI) and the Young Korean Medical Education \& Research Secretariat (Y-KOMERS) Medical Student Overseas Training Bootcamp Program, funded by the Ministry of Health \& Welfare, Republic of Korea, the KREONET Advanced Research Program Grant by the Korea Institute of Science and Technology Information (KISTI), the National Research Foundation of Korea (NRF) grant funded by the Ministry of Science and ICT, Republic of Korea (No.~RS-2026-NR121335), and infrastructure use from the Chameleon testbed, supported by NSF awards \#1419152, \#1743354, and \#2027170.
\end{acks}

\bibliographystyle{ACM-Reference-Format}
\bibliography{references}

\end{document}